\documentclass{svproc}
\usepackage{url}

\usepackage{amsmath,amssymb,amsfonts}
\usepackage{algorithmic} 
\usepackage{mathtools}
\usepackage{xfrac}	
\usepackage{siunitx}
\usepackage{graphicx}
\usepackage{multicol}
\usepackage{footmisc}
\usepackage[caption=false, font=footnotesize]{subfig}

\newcommand{\Res}[0]{\boldsymbol{\mathit{\delta}}}

\graphicspath{{./graphics/}}

\begin{document}
\mainmatter              
\title{Quantifying Uncertainties of Contact Classifications in a Human-Robot Collaboration with Parallel Robots}
\titlerunning{Quantifying Uncertainties in Contact Classification}  
%
\author{Aran Mohammad\inst{1}\and Hendrik Muscheid\and Moritz Schappler\and Thomas Seel}
\authorrunning{Aran Mohammad et al.} 
%
\tocauthor{Aran Mohammad, Moritz Schappler, Hendrik Muscheid, and Thomas Seel}
\institute{Leibniz University Hannover, Institute of Mechatronic Systems,\\30823 Garbsen, Germany,\\
\email{aran.mohammad@imes.uni-hannover.de},\\ WWW home page:
\texttt{https://www.imes.uni-hannover.de/en/}}

\maketitle              
\begin{abstract}
	In human-robot collaboration, unintentional physical contacts occur in the form of collisions and clamping, which must be detected and classified separately for a reaction. 
	If certain collision or clamping situations are misclassified, reactions might occur that make the true contact case more dangerous.
	This work analyzes data-driven modeling based on physically modeled features like estimated external forces for clamping and collision classification with a real parallel robot.
	The prediction reliability of a feedforward neural network is investigated. 
	Quantification of the classification uncertainty enables the distinction between safe versus unreliable classifications and optimal reactions like a retraction movement for collisions, structure opening for the clamping joint, and a fallback reaction in the form of a zero-g mode. 
	This hypothesis is tested with experimental data of clamping and collision cases by analyzing dangerous misclassifications and then reducing them by the proposed uncertainty quantification.
	Finally, it is investigated how the approach of this work influences correctly classified clamping and collision scenarios. 
\keywords{human-robot collaboration, data-driven modeling, parallel robots}
\end{abstract}
\section{Introduction}
	The human-robot collaboration allows an increase of performance for not fully-automated domains such as in assembly while simplifying human tasks. 
	Simultaneously, safety fences are no longer necessary, as implemented functions in the collaborative robots (cobots) are supposed to ensure safety.
	This can expand the available space in production capacity. 
	
	Due to safety requirements, limits on contact force and pressure must be maintained in the event of \emph{unintentional physical contacts} such as collision or clamping \cite{InternationalOrganizationforStandardization.2016}.
	Low moving masses of \emph{lightweight cobots} reduce the occurring contact forces. 
	However, the drives mounted on the joints increase the moving masses and the inertia of the kinematic chain. 
	An alternative approach is to use \emph{parallel robots} (PRs), where drives are typically fixed to the base and connected to a mobile platform via a kinematic chain \cite{Merlet.2006}. 
	The kinematic chains possess \emph{lower moving masses}, resulting in an increase in tolerable velocity while maintaining the same energy limits.

	Human safety must be ensured in a collaborative robot application --- either serial or parallel.
	Therefore, temporally transient and quasi-stationary \emph{contacts} (collision and clamping) between the human and the robot must be \emph{detected} and subsequently removed by a \emph{reaction} \cite{Haddadin.2017}. 
	\subsection{State of the Art}
		Contacts can be detected via \emph{exteroceptive information} using tactile \cite{Dahiya.2013} or visual measurements \cite{Rosenstrauch.2018,Merckaert.2022,Hoang.2022}. 
		\emph{Proprioceptive information} via built-in sensors in the robot offers advantages in terms of shorter sample times and lower hardware requirements.
		Physical methods enable \emph{contact detection, localization and identification} based on the measurement of joint angles and torques \cite{Kaneko.1994,Luca.2003,Haddadin.2017}.	
		Data-driven modeling is able to compensate drawbacks of physical models, since exact modeling is complex and modeling inaccuracies can occur. 
		This allows learning of the functional input-output-relationship directly using measured and also physically modeled data.
		 
		The \emph{contact detection} is realized in \cite{Popov.2017} as a classification task using a feedforward neural network (FFNN) to learn contact detection with the external joint torques of a serial robot as inputs. 
		To increase accuracy, temporal information can be integrated into a contact classifier by processing the time series of measured and physical quantities through convolutional neural networks \cite{Heo.2019,Park.2021,Zhang.2021}. 
		In \cite{Zhang.2021}, the uncertainty of contact detection is additionally quantified via the series of sequential prediction results.
		
		The \emph{isolation} (i.e. localization) of a collision can be learned for serial robots \cite{Popov.2017,Cioffi.2020,BriquetKerestedjian.2019} through data-based models with physically modeled features. 
		By incorporating physically modeled features such as the effects of contact forces in joint space and operational space, existing knowledge of kinematics and dynamics can be integrated into the learning of isolation. 
		In \cite{Popov.2017}, the determination of the contact location is achieved through an FFNN to predict a scalar quantity normalized to the entire length of the robot.
		The determination of the affected links is implemented in \cite{Cioffi.2020} by a support vector machine whose features are extracted via dimension reduction methods such as principal component analysis. 
		
		Data-driven \emph{discrimination} between collision and intentional interaction can be implemented via supervised machine learning methods with time series or features derived from contact dynamics \cite{Golz.2015,Zhang.2021}.
		The reason for using time series or physically modeled features is to learn about the temporal and dynamic effects of the functional input-output relationship.
		
		Methods of contact detection and reaction for an HRC with PRs exist in the authors' works. 
		The \emph{classification of the collided body} of a PR can be learned by physically modeled quantities concerning the orientation of the line of action in operational space \cite{Mohammad_IROS_Lokalisation.2023}. 		
		An FFNN \emph{distinguishing between a collision and clamping} at a PR is trained in \cite{Mohammad_IROS_Reaction.2023} and tested in different joint angle configurations.
		The estimated external forces represent the features that are processed to predict the two contact types.
		A second FFNN predicts the kinematic chain of clamping based on the location of the line of action of the external forces. 
	\subsection{Contributions}
		If certain collision or clamping situations are falsely classified, reactions might occur that make the true contact case more dangerous (Dangerous false classification and reaction - DFCR).
		In the DFCR cases, there is an optimal reaction for the prediction, but it leads to the increase of the contact forces of the actual contact.
		However, there is a suboptimal but safe fallback reaction (SFR), such as switching to zero-g mode.
		
		The approach of this work is an extension of the authors' work \cite{Mohammad_IROS_Reaction.2023} by quantifying clamping classification uncertainty to enable safe classifications and optimal reactions in form of a retraction movement and a structure opening. 
		If the prediction is unreliable, the suboptimal but safe fallback reaction in the form of a {zero-g} mode is performed.
		The zero-g mode is described as a suboptimal fallback reaction since it does not compensate for all robot dynamic effects in the contact case. 
		\begin{figure}[b!]
			\centering
			\includegraphics[width=1\columnwidth]{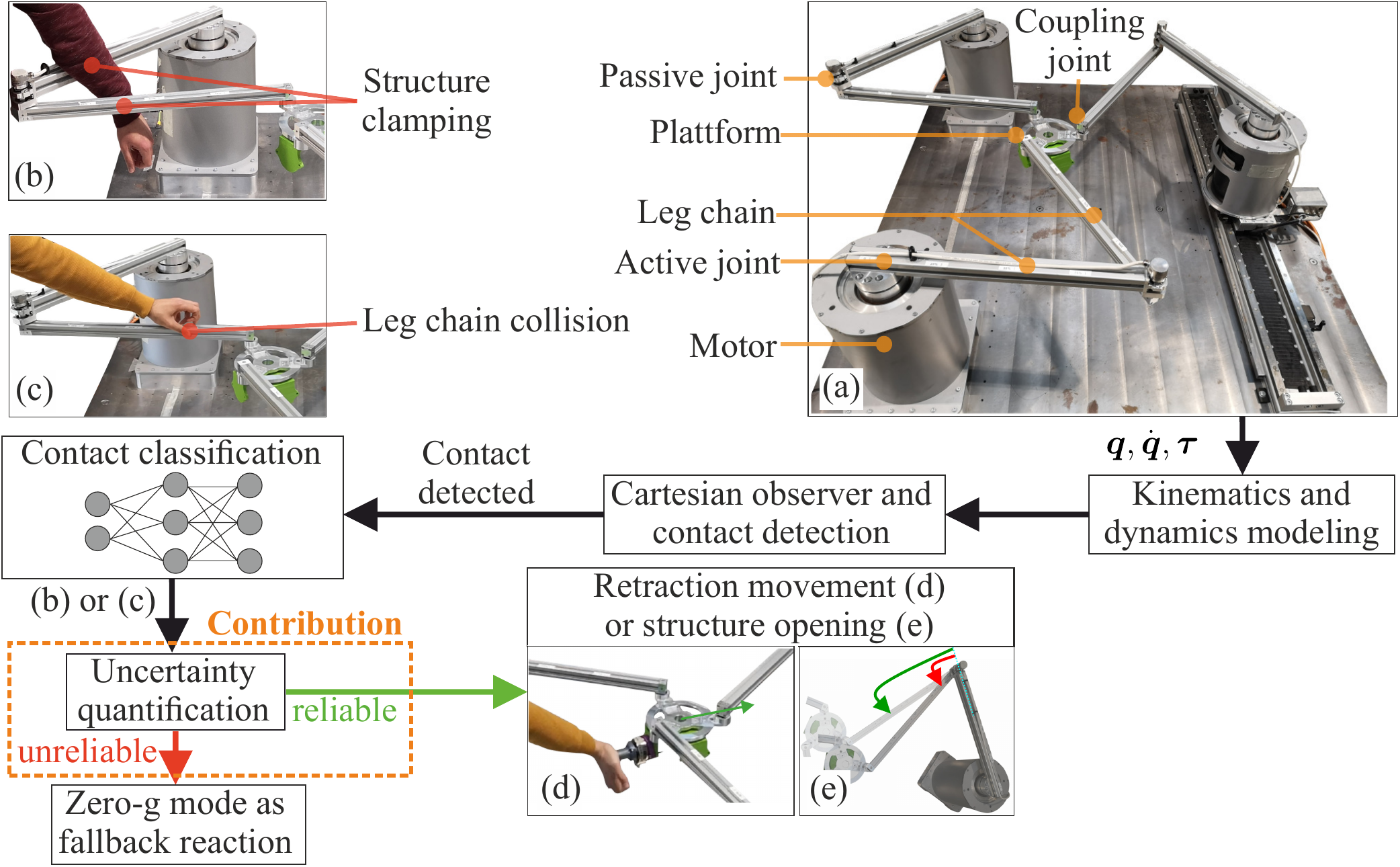}
			\caption{The parallel robot (a) is considered in this work with a clamping (b) and collision (c) contact. Compared to the related work, the contribution is a quantification of the contact classification uncertainty to enable safe classifications and optimal reactions (d),(e). If the uncertainty is large, the suboptimal but safe fallback reaction in the form of a zero-g mode is performed}
			\label{fig:titelbild_red}
			\vspace{0mm}
		\end{figure}		
		However, the zero-g mode reaction only requires the information of the contact detection, which is why it is not influenced by an incorrect clamping and collision classification and is also regarded as safe.
		Here, like the PR-related works \cite{Mohammad_IROS_Reaction.2023,Mohammad_IROS_Lokalisation.2023}, a classifier for clamping is tested in joint configurations that are unknown in training, which demonstrates a more realistic test scenario and allows an evaluation of the generalizability of the presented method. 
		As shown in Fig.~\ref{fig:titelbild_red}, monitoring the classification for sufficient accuracy and the avoidance of DFCRs lead to the contribution of this work:
	\begin{itemize}
		\item An approach to determine and distinguish dangerous false classifications and reactions (DFCRs) and less dangerous false classifications and reactions (LDFCRs) is presented using clamping and collision experiments of a real PR.
		\item A computationally efficient uncertainty measure from an FFNN allows for reducing the number of DFCRs by performing a safe fallback reaction (SFR) in the form of a zero-g mode.
		\item The effects of the uncertainty measure and SFR on false and true classifications are described and validated using real measured data.
	\end{itemize}
	The paper is structured as follows. 
	Starting with kinematics and dynamics modeling in Sect. \ref{sec:preliminaries}, the contact classification and uncertainty quantification are presented in Sect. \ref{sec:RelContactClass}. 
	Then, in Sect. \ref{sec:validation}, the PR used in this work is described, followed by an experimental evaluation of the reduced dangerous false classification and reactions. 
	Section \ref{sec:conlusions} concludes the paper.

\section{Preliminaries} \label{sec:preliminaries}
	This section is a brief summary of the work \cite{Mohammad_ICRA.2023} and begins with the description of the basics of kinematics (Sect.~\ref{ssec:kinematics}) and dynamics modeling (Sect.~\ref{ssec:dynamics}) of the used PR. 
	Subsequently, the disturbance observer (Sect.~\ref{ssec:observer}) is presented. 
	\subsection{Kinematics} \label{ssec:kinematics}
\begin{figure}[tb!]
			\centering
			\includegraphics[width=0.4\columnwidth]{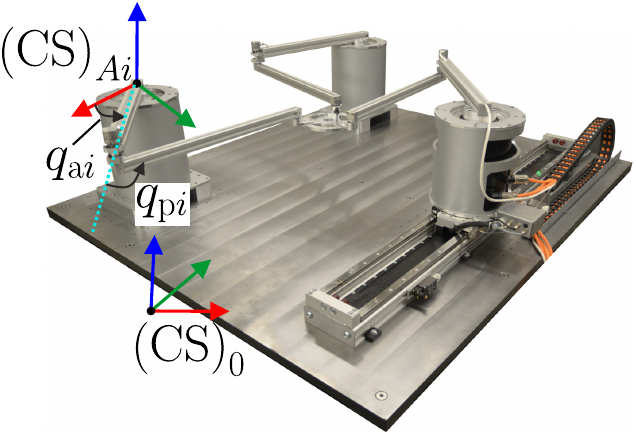}
			\caption{The 3-\underline{R}RR PR \cite{Mohammad_ICRA.2023} with the active and passive joint angles $q_{\mathrm{a}i},q_{\mathrm{p}i}$}
			\label{fig:3PRRR_real_skizze}
			\vspace{0mm}
		\end{figure}

		The following methods can be applied to any fully-parallel robot, but the modeling is described using the planar 3-\underline{R}RR parallel robot\footnotemark \footnotetext{The letter R denotes a revolute joint and underlining actuation. The actuated prismatic joint of the parallel robot is kept constant and is therefore not considered in the modeling.}
		shown in Fig.~\ref{fig:3PRRR_real_skizze} with $m{=}3$ platform degrees of freedom and $n{=}3$ leg chains \cite{Thanh.2012}. 
		Operational space coordinates (platform pose), active, passive, and coupling joint angles are represented respectively by $\boldsymbol{x}{\in}\mathbb{R}^m,\boldsymbol{q}_\mathrm{a}{\in}\mathbb{R}^n$ and $\boldsymbol{q}_\mathrm{p}, \boldsymbol{q}_\mathrm{c}{\in}\mathbb{R}^3$.
		
		The $n_i{=}3$ joint angles (active, passive, platform coupling) of each leg chain in $\boldsymbol{q}_i{\in}\mathbb{R}^{n_i}$ is stacked as $\boldsymbol{q}^\mathrm{T}{=}[\boldsymbol{q}_1^\mathrm{T}, \boldsymbol{q}_2^\mathrm{T}, \boldsymbol{q}_3^\mathrm{T}] {\in} \mathbb{R}^{3n}$. 
		By closing vector loops \cite{Merlet.2006} the kinematic constraints $\Res (\boldsymbol{q}, \boldsymbol{x}){=}\boldsymbol{0}$ are constructed. 
		Eliminating the passive joint angles $\boldsymbol{q}_\mathrm{p}$ yields the reduced kinematic constraints $\Res_\mathrm{red}(\boldsymbol{q}_\mathrm{a}, \boldsymbol{x}){=}\boldsymbol{0}$. 
		From this, the explicit analytic formulation of the active joint angles (inverse kinematics) can be calculated.
		Passive joint angles are measured to calculate an initial estimate of the platform's pose to encounter ambiguous forward kinematics. 
		The Newton-Raphson approach is then used since the encoders have different accuracies. 
		A time derivative of the kinematic constraints leads to 
		\begin{align}\label{eq_DifKin_Jac1}
			\dot{\boldsymbol{q}}&={-}\Res_{\partial \boldsymbol{q}}^{-1}\Res_{\partial \boldsymbol{x}} 	\dot{\boldsymbol{x}}=\boldsymbol{J}_{q,x}\dot{\boldsymbol{x}}\\
			\label{eq_DifKin_Jac2}
			\dot{\boldsymbol{x}}&= {-}\left(\Res_\mathrm{red}\right)_{\partial \boldsymbol{x}}^{-1} 	\left(\Res_\mathrm{red}\right)_{\partial \boldsymbol{q}_\mathrm{a}} \dot{\boldsymbol{q}}_\mathrm{a}=\boldsymbol{J}_{x,q_\mathrm{a}}\dot{\boldsymbol{q}}_\mathrm{a}
		\end{align}
		with the notation $\boldsymbol{a}_{\partial \boldsymbol{b}}{\coloneqq} \sfrac{\partial \boldsymbol{a}}{\partial \boldsymbol{b}}$ and the Jacobian matrices\footnotemark $\boldsymbol{J}_{q, x}{\in}\mathbb{R}^{\dim(\boldsymbol{q})\times m}$ and $\boldsymbol{J}_{x, q_\mathrm{a}}{\in}\mathbb{R}^{m\times n}$.\footnotetext{For the sake of readability, dependency on $\boldsymbol{q}$ and $\boldsymbol{x}$ is omitted.}
	\subsection{Dynamics} \label{ssec:dynamics}
		The Lagrangian equations of the second kind, the \textit{subsystem} and \textit{coordinate partitioning} methods provide the equations of motion in the operational space to eliminate the constraint forces \cite{Thanh.2009}.
		The inverse dynamics equation 
		\begin{equation} \label{eq_dyn}
			\boldsymbol{M}_x \ddot{\boldsymbol{x}}{+} \boldsymbol{c}_x {+} \boldsymbol{g}_x{+} 	\boldsymbol{F}_{\mathrm{fr},x}= \boldsymbol{F}_\mathrm{m} {+} \boldsymbol{F}_\mathrm{ext}
		\end{equation}
		applies for the present PR with the following generalized forces $\boldsymbol{F}{\in}\mathbb{R}^m$ (also including moments) acting on the platform. 
		Equation \ref{eq_dyn} consists of $\boldsymbol{M}_x$ as the symmetric positive-definite inertia matrix, $\boldsymbol{c}_x{=}\boldsymbol{C}_x\dot{\boldsymbol{x}}$ as the vector/matrix of the centrifugal and Coriolis terms, $\boldsymbol{g}_x$ as the gravitational effects (in the {non-planar} case), $\boldsymbol{F}_{\mathrm{fr},x}$ as the viscous and Coulomb friction components, $\boldsymbol{F}_\mathrm{m}$ as the forces based on the motor torques and $\boldsymbol{F}_{\mathrm{ext}}$ as external forces. 
		The forces $\boldsymbol{F}_\mathrm{m}$ are projected into the actuated joint coordinates by the principle of virtual work $\boldsymbol{\tau}_\mathrm{a}{=}\boldsymbol{J}_{x,q_\mathrm{a}}^\mathrm{T}\boldsymbol{F}_\mathrm{m}$. 
	\subsection{Generalized-Momentum Observer} \label{ssec:observer}
		Introduced by \cite{Luca.2003}, a residual of the generalized momentum $\boldsymbol{p}_x{=}\boldsymbol{M}_x \dot{\boldsymbol{x}}$ is set up in this work in the operational space coordinate $\boldsymbol{x}$ as the minimal coordinate for the dynamics of PRs. 
		The derivative of the residual w.r.t. time is $\mathrm{d}(\hat{\boldsymbol{F}}_\mathrm{ext}) / \mathrm{dt} {=} \boldsymbol{K}_{\mathrm{o}} (\dot{\boldsymbol{p}}_x {-} \dot{\hat{\boldsymbol{p}}}_x )$ with $\boldsymbol{K}_\mathrm{o}{=}\mathrm{diag}(k_{\mathrm{o},1},k_{\mathrm{o},2}, \dots ,k_{\mathrm{o},m}){>}\boldsymbol{0}$ as the observer gain matrix \cite{Luca.2003}. 
		Transforming (\ref{eq_dyn}) to $\hat{\boldsymbol{M}}_x\ddot{\boldsymbol{x}}$ and substituting it into the integral of $\dot{\hat{\boldsymbol{F}}}_\mathrm{ext}$ over time, leads to 
		\begin{align} \label{eq:mo}
			\hat{\boldsymbol{F}}_\mathrm{ext} &= \boldsymbol{K}_\mathrm{o} \left( \hat{\boldsymbol{M}}_x 	\dot{\boldsymbol{x}} {-} \int_{0}^t \boldsymbol{F}_\mathrm{m} {-} \hat{\boldsymbol{\beta}} {+} \hat{\boldsymbol{F}}_\mathrm{ext} \mathrm{d}\tilde{t} \right) \;, \\ 
			\nonumber
			\hat{\boldsymbol{\beta}} &= \hat{\boldsymbol{g}}_x {+} \hat{\boldsymbol{F}}_{\mathrm{fr},x} 	{+} \left( \hat{\boldsymbol{C}}_x {-}\dot{\hat{\boldsymbol{M}}}_x \right)\dot{\boldsymbol{x}} = \hat{\boldsymbol{g}}_x {+} \hat{\boldsymbol{F}}_{\mathrm{fr},x}{-}\hat{\boldsymbol{C}}_x^\mathrm{T} \dot{\boldsymbol{x}}
		\end{align}
		with $\dot{\hat{\boldsymbol{M}}}_x{=}\hat{\boldsymbol{C}}_x^\mathrm{T}{+}\hat{\boldsymbol{C}}_x$ \cite{Haddadin.2017,Ott.2008}. 
		Under the condition $\hat{\boldsymbol{\beta}}{\approx}\boldsymbol{\beta}$ it follows 
		\begin{align}
			\boldsymbol{K}_\mathrm{o}^{-1} \dot{\hat{\boldsymbol{F}}}_\mathrm{ext} {+} 	\hat{\boldsymbol{F}}_\mathrm{ext}{=}\boldsymbol{F}_\mathrm{ext} \;,
		\end{align}		 
		which corresponds to a linear and decoupled error dynamics of the generalized-momentum observer, exponentially converging to the external force projected to platform coordinates. 
\section{Reliable Contact Classification} \label{sec:RelContactClass}
	An FFNN for clamping and collision classification is presented at the beginning of this section in \ref{ssec:ffnn_class}. 
	The dataset (Sect.~\ref{ssec:dfcr}) and the approach for uncertainty quantification (Sect.~\ref{ssec:Quant_Uncert}) are described afterward.
	\subsection{Clamping and Collision Classification of a Neural Network} \label{ssec:ffnn_class}
		An FFNN performs a binary classification of a contact into the output classes $\{$Collision, Clamping$\}$ based on the input data $\hat{\boldsymbol{F}}_\mathrm{ext}$ from~(\ref{eq:mo}) cf. \cite{Mohammad_IROS_Reaction.2023}.
		The FFNN is trained using the gradient-based optimization method Adam \cite{Kingma.22122014,FabianPedregosa.2011}. 
		The hyperbolic tangent is chosen as the nonlinear activation function. 
		To avoid underfitting and overfitting, a hyperparameter optimization of the $L_2$ regularization term, as well as of the network structure with the number of neurons and hidden layers is performed in terms of a grid search.
	\subsection{Dangerous False Classifications and Reactions} \label{ssec:dfcr}
		\begin{figure}[bt!]
			\centering
			\includegraphics[width=1\columnwidth]{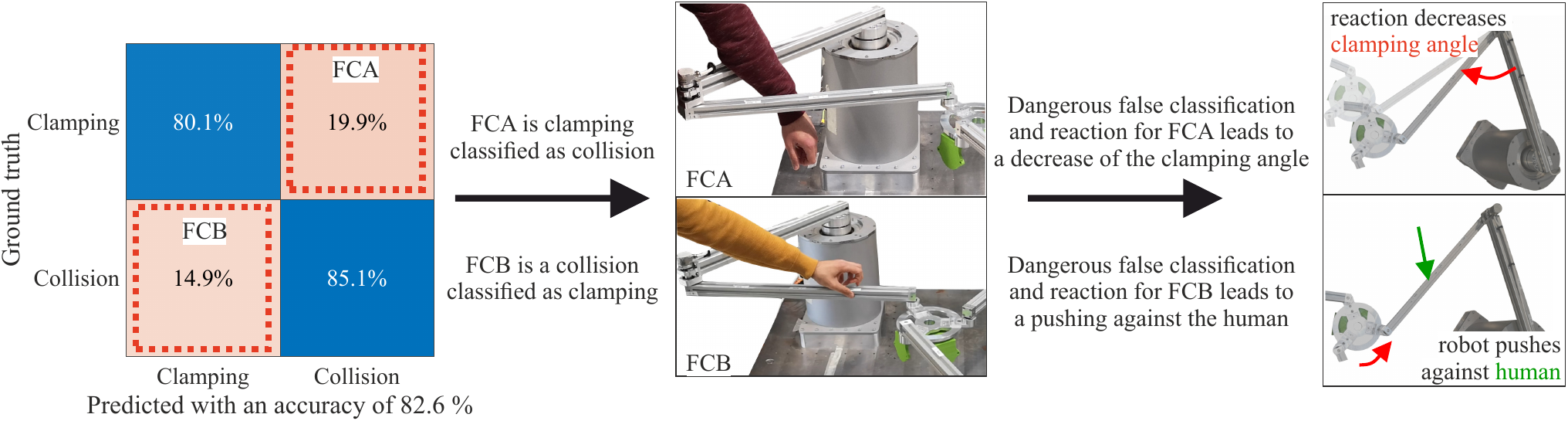}
			\caption{Qualitative demonstration for determination and distinguishing dangerous false classification and reactions}
			\label{fig:Falschklassifikation_red}
			\vspace{0mm}
		\end{figure}
		Figure~\ref{fig:Falschklassifikation_red} shows the approach to determine DFCRs starting from misclassified collisions and clamping from \cite{Mohammad_IROS_Reaction.2023}. 
		Measurements of three joint angle configurations are available, each with three clamping cases and seven collisions (six links and the mobile platform). 
		The dataset consists of 80k measurement samples. 
		To evaluate the generalization of the FFNN, training and testing are performed with data from different joint angle configurations of the PR. 
		
		The off-diagonal entries of the row-normalized confusion matrix in Fig.~\ref{fig:Falschklassifikation_red} are termed as false classifications A and B (FCA/FCB). 
		If a clamping is falsely classified and the retraction direction of the reaction decreases the clamping angle and raises the contact forces, the data point is labeled as DFCR.
		
		Similarly, if a collision is misclassified and the robot structure presses against the human as a result of the structure opening of the falsely classified clamping, then the contact forces increase. 
		This data point is labeled as DFCR.
		Misclassified cases without an increase in contact force are termed as less dangerous false classification and reaction (LDFCR).
	\subsection{Quantifying Unreliable Classifications} \label{ssec:Quant_Uncert}
		The neuron outputs in the output layer of the FFNN are normalized to the value range between 0 and 1 using a softmax or logistic unit function. 
		These values are now considered as probabilities to allow a reduction of misclassification. 
		Classifications with a probability less than a limit $p_\mathrm{th}$ are assumed to be DFCR and the SFR is performed.
		Another case is a correct contact classification with a large uncertainty, so the SFR is performed instead of the optimal response. 

\section{Validation} \label{sec:validation}
	After presenting the experimental setup with the uncertainty quantification (Sect.~\ref{ssec:expSetup_UncQuant}), the reduction of dangerous false classified clamping contacts (Sect.~\ref{ssec:Clamp_as_Coll_class}) and collisions (Sect. \ref{ssec:Coll_as_Clamp_class}) are considered.
	Next, the results of correct contact classifications (Sect. \ref{ssec:UncQuant_Infl_CorrClass}) are presented.
	\subsection{Experimental Setup with Uncertainty Quantification} \label{ssec:expSetup_UncQuant}
\begin{figure}[bt!]
			\centering
			\includegraphics[width=1\columnwidth]{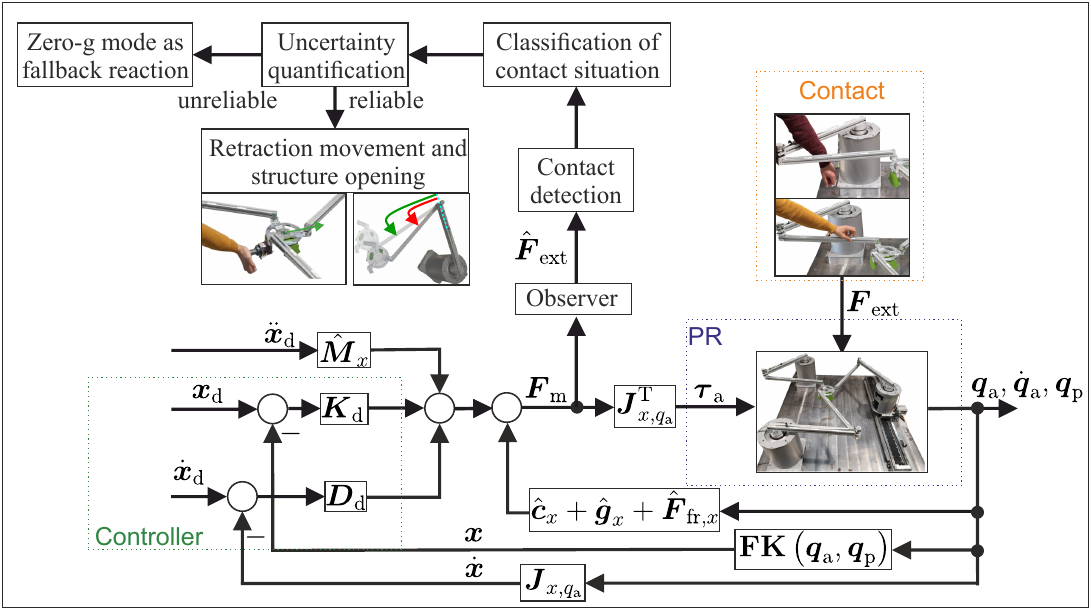}
			\caption{Block diagram with the experimental setup --- The uncertainty quantification extends the contact reactions from \cite{Mohammad_IROS_Reaction.2023}}
			\label{fig:Ablauf_3PRRR_red}
			\vspace{0mm}
		\end{figure}

		Figure~\ref{fig:Ablauf_3PRRR_red} represents the block diagram of the system which is operated at a sampling rate of $\SI{1}{\kilo\hertz}$. 
		The parameterization of a Cartesian impedance controller \cite{Taghirad.2013,Ott.2008} is set to $\boldsymbol{K}_\mathrm{d}{=}\mathrm{diag}(\SI{2}{\newton/ \milli\meter}, \SI{2}{\newton/\milli\meter}, \SI{85}{Nm/\radian})$. 
		Since the direct drives are used without gears and associated friction, joint torque control via the motor current is permissible. 
		Critical damping is achieved using the factorization damping design \cite{AlbuSchaffer.2003}. 
		The dynamics base parameters from \cite{Thanh.2012} are used and symmetric dynamics properties of the three leg chains are assumed except the friction. 
		The observer is parameterized with $k_{\mathrm{o},i}{=}\frac{1}{\SI{50}{\milli \second}}$.
		More detailed information on the experimental setup can be found in \cite{Mohammad_ICRA.2023}.
	
		Due to the deviations between observed and measured external force, thresholds for contact detection are defined empirically by $\boldsymbol{\epsilon}_\mathrm{ext}^\mathrm{T}{=}[\SI{10}{\newton},\SI{10}{\newton},\SI{1}{Nm}]$. 
		As soon as $|\hat{F}_{\mathrm{ext},i}|{>}\epsilon_{\mathrm{ext},i}$, the clamping and collision classification is initiated. 
		All the following results are based on data fulfilling the condition $|\hat{F}_{\mathrm{ext},i}|{>}\epsilon_{\mathrm{ext},i}$.
		A reaction in the form of a retraction and a structural opening depending on the clamping classification is only executed as soon as the probability $p$ of the classification is larger than the threshold $p_\mathrm{th}$. 
		If $p{\le}p_\mathrm{th}$ holds, zero-g mode is activated as a fallback reaction.
		The rates of DFCR/LDFCR from the FCA/FCB cases are determined with the varied threshold $p_\mathrm{th}$.		
		The ratio of SFR in the correctly classified clamping and collisions is calculated with the varying threshold $p_\mathrm{th}$.
	\subsection{False Classification A: Clamping Classified as Collision} \label{ssec:Clamp_as_Coll_class}
		\begin{figure}[tb!]
			\centering
			\includegraphics[width=1\columnwidth]{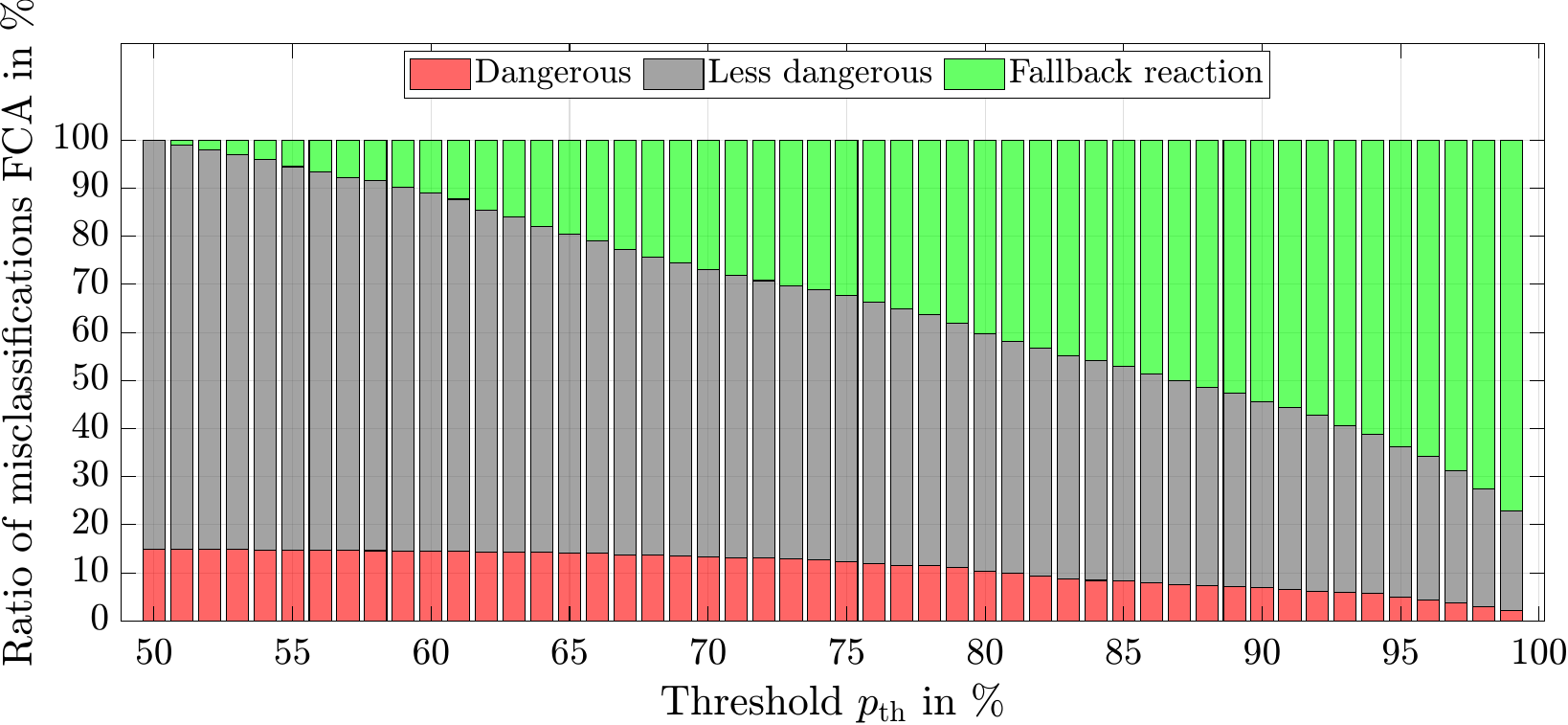}
			\caption{Ratios of the three reaction scenarios after false classification A (FCA --- Clamping is classified as collision) and for varying $p_\mathrm{th}$}
			\label{fig:Classification_Clamping_PropMisclFCA_over_threshold}
			\vspace{0mm}
		\end{figure}
		Figure~\ref{fig:Classification_Clamping_PropMisclFCA_over_threshold} shows the ratios of reaction scenarios for clamping contacts classified as collisions over the threshold $50\%{\le}p_\mathrm{th}{\le}99\%$. 
		
		This range is chosen because uncertainty quantification only affects the reaction process in this range due to the binary output of the clamping classification. 
		Thus, the probability of the output for the binary classification is $p{\ge} 50\%$.
		It can be seen at $p_\mathrm{th}{=}50\%$ that $15\%$ of FCA belong to DFCR. 
		The nonlinear influence of $p_\mathrm{th}$ on the presented ratios is recognizable. 
		Thus, for $p_\mathrm{th}{=}75\%$ the ratio of DFCR is reduced to only $13\%$. 		
		Setting $p_\mathrm{th}{=}99\%$, the ratio of DFCR (LDFCR) is reduced to $3\%$ ($20\%$). 
		This is realized by the SFR, which is now executed in a total of $77\%$ of the FCA cases.

	\subsection{False Classification B: Collision Classified as Clamping} \label{ssec:Coll_as_Clamp_class}
		\begin{figure}[b!]
			\centering
			\includegraphics[width=1\columnwidth]{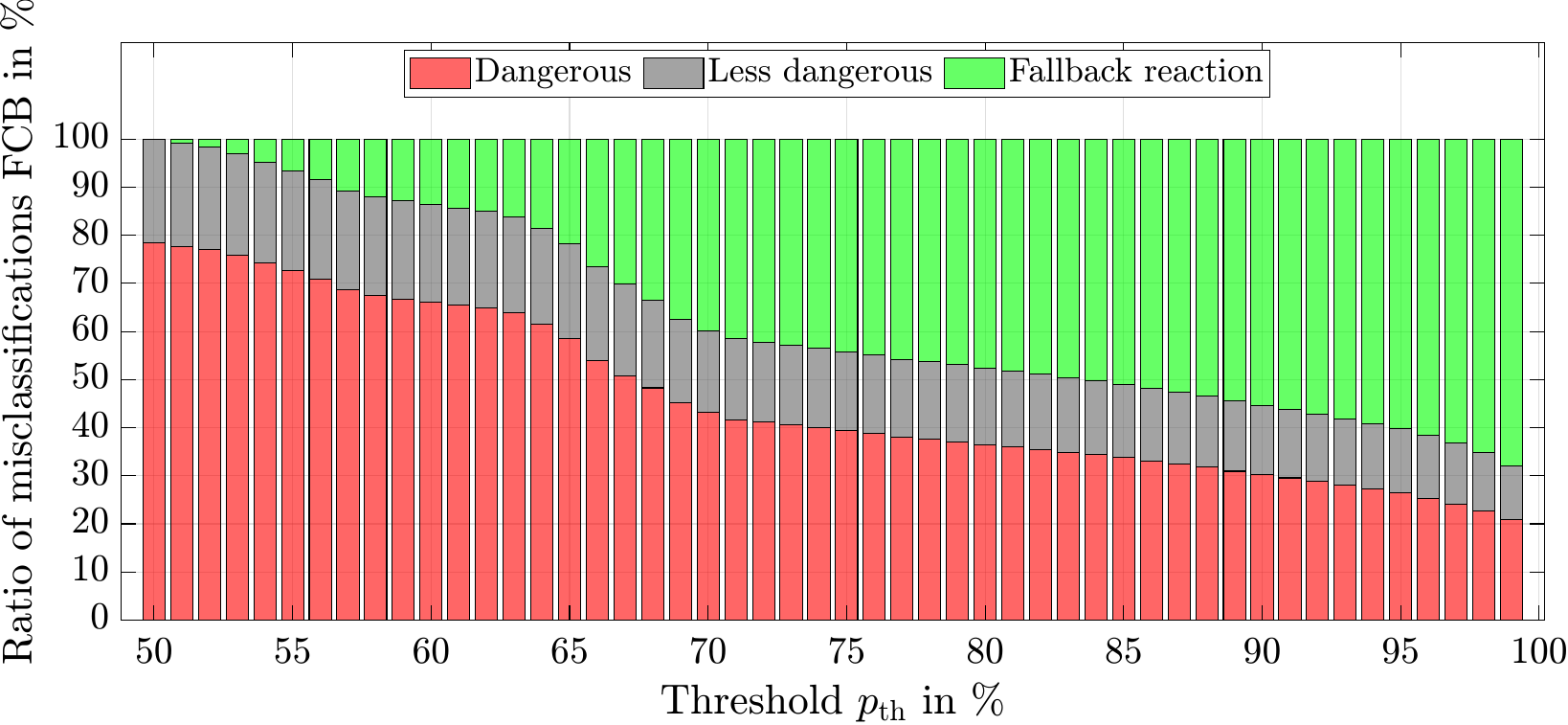}
			\caption{Ratios of the three reaction scenarios after false classification B (FCB --- Collision is classified as clamping) and for varying $p_\mathrm{th}$}
			\label{fig:Classification_Clamping_PropMisclFCB_over_threshold}
			\vspace{0mm}
		\end{figure}
		The ratios of the different reactions concerning FCB is depicted in Fig.~\ref{fig:Classification_Clamping_PropMisclFCB_over_threshold}.
		Noticeable in $p_\mathrm{th}{=}50\%$ is the high ratio $78\%$ of DFCR for collisions which are classified as a clamping. 
		In these cases, a structural opening leads to the falsely assumed removal of the clamping, but results in an increase of the true contact forces of the collision.
		DFCRs occur at $p_\mathrm{th}{=}99\%$ only in $21\%$ of the cases.
		In contrast to the previous FCA, LDFCRs occur in $p_\mathrm{th}{=}50\%$ ($p_\mathrm{th}{=}99\%$) only in $22\%$ ($11\%$) of the cases. 
		Thus, zero-g mode is performed as SFR in up to $68\%$ of FCB. 
		
	\subsection{Influence of the Uncertainty Quantification on Correctly Contact Classification} \label{ssec:UncQuant_Infl_CorrClass}
\begin{figure}[t!]
			\centering
			\includegraphics[width=1\columnwidth]{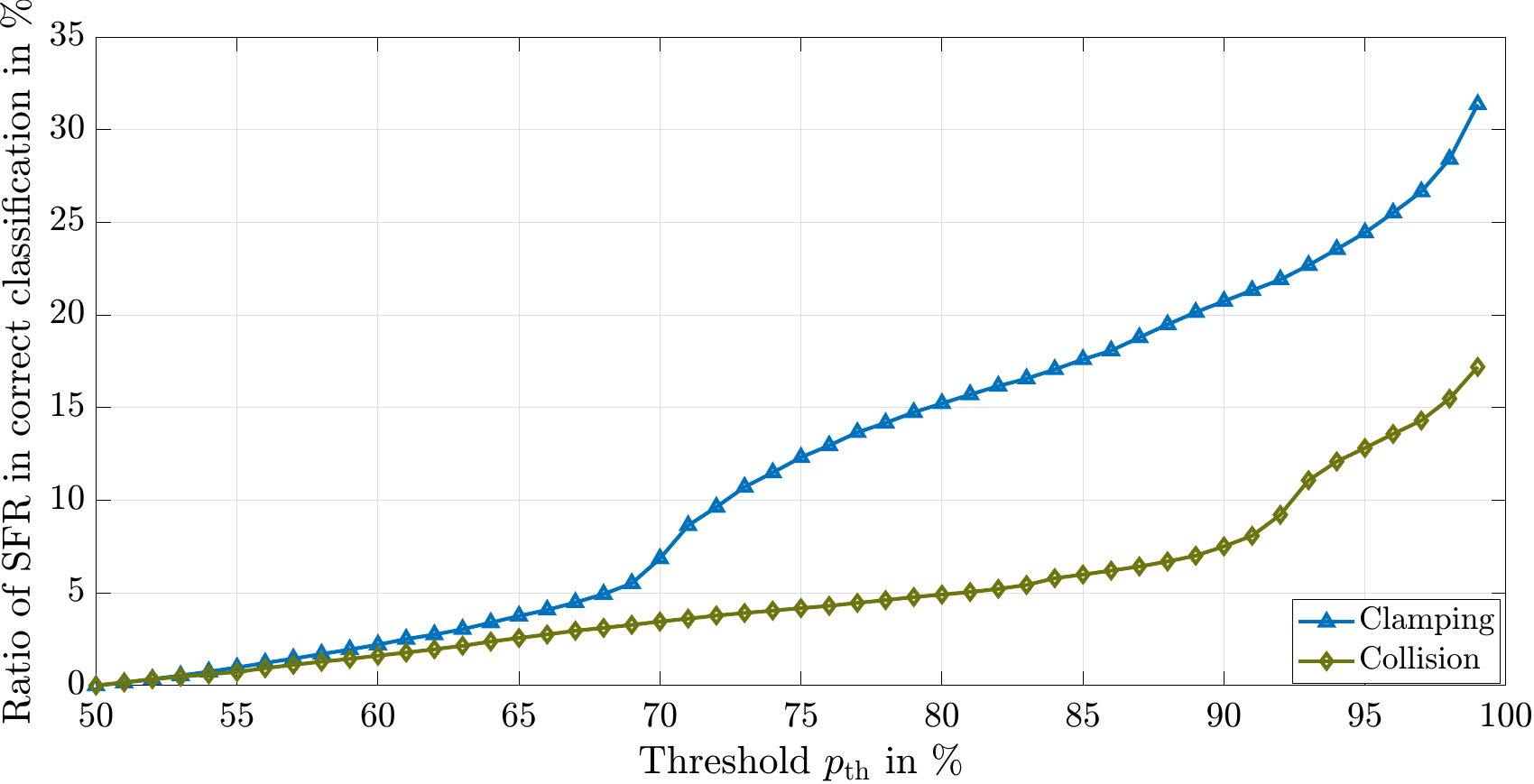}
			\caption{Ratios of the safe fallback reaction in the correctly classified clamping and collision cases for varying $p_\mathrm{th}$}
			\label{fig:Classification_Clamping_PercCorrClass_over_threshold}
		\vspace{0mm}
		\end{figure}

		The previous results relate only to misclassifications. 
		Since the uncertainty quantification and reaction selection also affect correctly classified contacts, these results are described using the ratios in Fig.~\ref{fig:Classification_Clamping_PercCorrClass_over_threshold}.
		It is noticeable that the correctly classified collision and clamping cases are nonlinear and affected by the SFR to different extents. 
		Especially for clamping contacts, the ratio of SFR is increased significantly larger when $p_\mathrm{th}{>}70\%$ is chosen. 
		For $p_\mathrm{th}{=}99\%$, the uncertainty quantification leads to SFR occurring in up to $28\%$ ($16\%$) cases in the correctly classified cases of clamping (collisions). 
		However, these scenarios are not considered critical because the change to zero-g mode is also assumed to be a safe response.
				
		The results show that uncertainty quantification correlates with the number of misclassified clamping and collision contacts. 
		In addition, the threshold can be used to parameterize the switching of the reactions, which leads to the ratios of correct and false classifications shown in this section.

\section{Conclusions} \label{sec:conlusions}
	In a human-robot collaboration (HRC) with parallel robots (PRs), the probability of clamping and collision situations with a falsely predicted contact type and resulting reaction that could aggravate the contact case is reduced from $15\%$ to $3\%$ and $78\%$ to $21\%$ using the simple but effective approach of the work. 
	This is achieved by a less frequent optimal reaction regarding the predicted contact type.
	However, this condition of the reaction causes the execution of the also safe fallback reaction instead of the optimal response to clamping and collision cases in $28\%$ and $16\%$ of the correct classifications. 
	Improving this trade-off in a larger reduction of dangerous misclassifications with increased ratios of optimal responses in correctly classified cases remains future work. 
	Moreover, the results show that the proposed method can contribute to a safe HRC with PRs even in joint angle configurations that were unknown during training. 
	This is an important step towards data-driven methods performing more reliably and safely for real applications. 
%
%
	\bibliographystyle{spmpsci}
	\bibliography{literatur}
\end{document}